\documentclass[10pt,twocolumn,letterpaper]{article}

\usepackage[pagenumbers]{cvpr} %

\usepackage[dvipsnames]{xcolor}

\definecolor{cvprblue}{rgb}{0.21,0.49,0.74}
\usepackage[pagebackref,breaklinks,colorlinks,citecolor=cvprblue]{hyperref}

\usepackage{bm}

\usepackage{algorithm}
\usepackage{algpseudocode}

\newfloat{afloat}{t!}{log}
\floatname{afloat}{Iterative Update}
\DeclareCaptionLabelFormat{unnumbered}{#1}
\newfloat{bfloat}{t!}{log}
\floatname{bfloat}{Single-Pass Update}
\DeclareCaptionLabelFormat{unnumbered}{#1}
\algrenewcommand\algorithmicindent{0.75em}
\usepackage[dvipsnames]{xcolor}
\usepackage{array}
\newcommand{\PreserveBackslash}[1]{\let\temp=\\#1\let\\=\temp}

\def\paperID{1026} %
\def\confName{CVPR}
\def\confYear{2024}

\title{Efficient-NeRF2NeRF: Streamlining Text-Driven 3D Editing with Multiview Correspondence-Enhanced Diffusion Models}

\author{
Liangchen Song$^1$, Liangliang Cao$^1$, Jiatao Gu$^1$, Yifan Jiang$^{12}$, Junsong Yuan$^3$, Hao Tang$^1$\\
$^1$Apple \quad $^2$UT Austin \quad $^3$University at Buffalo
}

\begin{document}

\twocolumn[{%
\renewcommand\twocolumn[1][]{#1}%
\maketitle
\begin{center}
    \centering
    \captionsetup{type=figure}
    \includegraphics[width=\textwidth,trim={0 0 0 0}]{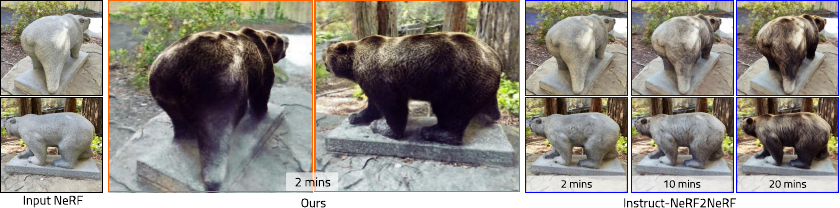}
    \captionof{figure}{\textbf{Efficient NeRF editing within 2 minutes}. 
    We present a framework that aims to enhance the efficiency of editing NeRF models using text-based instructions. The key factor contributing to this efficiency is our regularized diffusion scheme, which enables the direct generation of multiview-consistent images.
    (Prompt: \textit{``Turn the bear into a grizzly bear''}.)}
\end{center}%
}]

\begin{abstract}
The advancement of text-driven 3D content editing has been blessed by the progress from 2D generative diffusion models. However, a major obstacle hindering the widespread adoption of 3D content editing is its time-intensive processing. This challenge arises from the iterative and refining steps required to achieve consistent 3D outputs from 2D image-based generative models. 
Recent state-of-the-art methods typically require optimization time ranging from tens of minutes to several hours to edit a 3D scene using a single GPU. 
In this work, we propose that by incorporating correspondence regularization into diffusion models, the process of 3D editing can be significantly accelerated. This approach is inspired by the notion that the estimated samples during diffusion should be multiview-consistent during the diffusion generation process. By leveraging this multiview consistency, we can edit 3D content at a much faster speed. In most scenarios, our proposed technique brings a 10$\times$ speed-up compared to the baseline method and completes the editing of a 3D scene in 2 minutes with comparable quality.
Project page: \href{https://lsongx.github.io/projects/en2n.html}{https://lsongx.github.io/projects/en2n.html}.
\end{abstract}

\section{Introduction}
The recent accomplishments in foundational 2D editing methods \cite{Bommasani2021FoundationModels,rombach2022high,brooks2023instructpix2pix} have enabled us to personalize and modify 3D captured scenes \cite{instructnerf2023}, which hold great appeal and significant practical value. However, despite the impressive results obtained through these recent developments, the existing 3D editing methods \cite{sella2023vox,wang2023nerf,instructnerf2023} still suffer from the drawback of prolonged optimization duration, which often takes tens of minutes.

Achieving efficient 3D editing using text-driven 2D image editors presents a challenging task. This difficulty primarily arises from the fact that text-driven image editors typically operate on a per-image basis. Consequently, applying these editors to multiview images often results in ineffective correspondence among views, hence the edited 2D images cannot be directly leveraged for updating 3D content. 
Although significant advancements have been made in enhancing multiview correspondence, such as score-distillation sampling \cite{poole2022dreamfusion}, existing approaches frequently necessitate retraining and backpropagation, thereby making multiview editing inefficient.

\begin{figure}
\scriptsize
    \centering
    \includegraphics[width=\columnwidth]{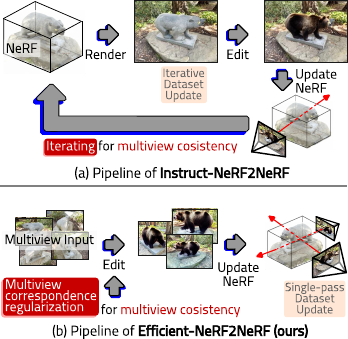}
    \caption{\textbf{Comparison of editing frameworks.} 
    Our approach significantly enhances the speed of the editing process by directly updating the NeRF using edited multiview images. In contrast, the SoTA method, Instruct-NeRF2NeRF, needs to iteratively edit and re-render images.
    }
    \label{fig:teaser-compare}
\end{figure}

This paper proposes an approach wherein we apply regularization to the diffusion denoising process to enhance its multiview correspondence. 
Performing direct edits on a collection of multi-view images leads to efficient 3D editing capabilities compared to existing works like Instruct-NeRF2NeRF \cite{instructnerf2023} (see \cref{fig:teaser-compare}). 
This is because the iterative updating framework \cite{instructnerf2023} can now be simplified by treating the process as a text-driven editing of 2D images and then updating the 3D representation.

Our key insight is that in 3D editing, where multiple-view images serve as inputs, we benefit from the observations of the existing correspondences between these images. 
If we can progressively rectify the inconsistencies observed across multiple views
during the diffusion sampling process,  we can effectively generate a collection of images that exhibit strong multiview consistency and subsequently update the 3D contents in a much more efficient way. Note that the idea of multiview consistency is also used in 3D generation tasks like text-to-3D \cite{poole2022dreamfusion} or image-to-3D \cite{xu2023neurallift} where correspondence needs to be generated.

However, ensuring consistent prediction across various perspectives is a challenging problem due to two reasons. Firstly, the color of a 3D point may vary depending on the viewing angle. Consequently, naively imposing identical estimations leads to unrealistic outcomes. Secondly, prevalent 2d diffusion models denoise in the latent space, where each latent vector represents an image patch. However, such latent representation cannot best capture the variations of image patches across different viewpoints. 

We address the aforementioned challenges through a two-fold approach: mitigating the impact of regularization for diffusion and minimizing the influence of inconsistent edits for training radiance fields. To tackle the regularization issue, we draw inspiration from recent studies that reveal how the diffusion process prioritizes overall image structure in the early denoising stages while focusing on texture details in the later stages \cite{meng2021sdedit}. Based on this insight, we propose a regularization strategy that places more emphasis on reducing inconsistency during the early steps of diffusion, and gradually removes the regularization towards the later steps. By softening the regularization, we can generate visually appealing images. However, the generated images may still not strictly consistent across multiviews. To overcome this, we adopt a loss function that encourages the 3D content optimization process to align the style of image patches using the Gram matrix \cite{gatys2016image} and randomly switch between the photometric loss and the style loss. %
In our method, this loss function guides the edited 3D to minimize the impact of multiview inconsistency outputs, such as shifts or deformations.
To sum up, our contributions are as follows:
\begin{itemize}
    \item 
    This paper proposes a text-driven 3D content editing framework, utilizing the power of 2D image editing techniques. This framework grants users the ability to efficiently edit 3D content, 10 times faster than 
    current radiance field editing techniques such as Instruct-NeRF2NeRF \cite{instructnerf2023}, which may require approximately 20 minutes.
    \item 
    This paper develops a regularization technique to preserve the multiview correspondence across a collection of images, eliminating the requirements for retraining the diffusion network. Moreover, to mitigate the impact of inconsistent generation on the tuning of the radiance field, we propose the incorporation of a style matching loss.
\end{itemize}

\section{Related Works}
\paragraph{Generating 3D Contents with Foundational 2D Generative Models.}
3D editing can be conceptualized as a conditioned generation challenge. In recent years, foundational 2D generative models \cite{Bommasani2021FoundationModels} have showcased remarkable prowess in open-set generation \cite{ramesh2022hierarchical,rombach2022high,saharia2022photorealistic}. The utilization of these 2D generative foundational models for open-set 3D content generation has garnered significant attention.
DreamFusion \cite{jain2022zero,poole2022dreamfusion} proposed Score Distillation Sampling (SDS) based on probability density distillation to get 3D generation priors from 2D generative models. 
Score Jacobian Chaining \cite{wang2023score} applied chain rule on the learned gradients and back-propagate the score of a 2D diffusion model through the Jacobian of a differentiable renderer for 3D content.
Given that 2D images are conventionally generated on a view-by-view basis, the central quandary in this avenue of research revolves around the identification of a multifaceted supervisory signal that ensures 3D consistency from the purview of 2D models, which can be improved in various aspects such as on the 2D generation conditioning \cite{armandpour2023re,chen2023fantasia3d,chen2023it3d,huang2023dreamtime,seo2023ditto,wang2023prolificdreamer,wu2023hd,zhu2023hifa,qian2023magic123,lin2023consistent123} and on the underlying 3D representation \cite{lin2023magic3d,yu2023points,tang2023dreamgaussian,chen2023text,seo2023let,tsalicoglou2023textmesh}. Besides pure text as conditioning, some methods are developed to take a single image as the input \cite{melas2023realfusion,raj2023dreambooth3d,shen2023anything,tang2023make,xu2023neurallift,xu2022sinnerf,qian2023magic123}.
The aforementioned methods primarily tackle a common challenge: the task of updating 3D representations in the presence of potentially multiview inconsistent images.
Within our approach, we incorporate a style loss component to mitigate the issue of multiview inconsistency. This strategy is employed with the specific aim of facilitating 3D editing, which diverges notably from the task of generating content from scratch.

\paragraph{Multiview Consistent Image Generation.}
Since multiview consistency stands as the cornerstone in producing high-quality 3D content, several methods are dedicated to refining multiview consistency in the process of 2D image generation. A prevalent research direction posits the augmentation of 2D image generation models with a heightened awareness of the imaging process, such as camera poses.
3DiM \cite{watson2022novel} proposed to use camera poses as a conditioning input to the denoising model in diffusion. Zero-1-to-3 \cite{liu2023zero} further improved this line of work with foundational 2D generative models and effective advanced multiview consistency modeling. One-2-3-45 \cite{liu2023one} further accelerated the sampling and reconstruction process of Zero-1-to-3. 
Instead of modeling the view-conditioned distribution of images, Viewset Diffusion \cite{szymanowicz2023viewsetdiffusion} and SyncDreamer \cite{liu2023syncdreamer} proposed to model the distribution of multiview images directly.
MVDiffusion \cite{tang2023mvdiffusion} and Consistent-1-to-3 \cite{ye2023consistent1to3} used epipolar geometry based attention across the views to improve multiview consistency, while MVDream \cite{shi2023mvdream} demonstrated the effectiveness of directly using the self-attention in Stable Diffusion \cite{rombach2022high}.
Our research distinguishes itself from the methodologies above by focusing on 3D editing, wherein we presuppose correspondence as an input to the generative process. In contrast, the previously mentioned approaches necessitate the generation of correspondence due to the absence of 3D inputs.

\paragraph{Radiance Fields Editing.}
Radiance Fields \cite{adelson1991plenoptic} serve as a means for representing 3D content, and they can be parameterized through neural networks \cite{mildenhall2020nerf,instantngp}, multiplane images \cite{zhou2018stereo,wizadwongsa2021nex}, point clouds \cite{xu2022point,kerbl20233d}, and others \cite{PlenOctrees,yu_and_fridovichkeil2021plenoxels,dvgo,chen2023neurbf,tensorf}.
Since a radiance field represents a 3D scene, editing it can be achieved with human-designed graphics knowledge, such as volume geometry \cite{ZhangLYZZWZXY21,yuan2022nerf,li2023climatenerf,yuan2023interactive,qiao2023dynamic,tschernezki2022neural,yang2021geometry}, color \cite{kuang2023palettenerf,gong2023recolornerf,lee2023ice} and lighting \cite{boss2021nerd,srinivasan2021nerv,boss2021neuralpil}. 
Some works \cite{chiang2022stylizing,huang2021learning,huang2022stylizednerf,fan2022unified,nguyen2022snerf,zhang2022arf} proposed to stylize a radiance field with 2D image style adaptors \cite{gatys2016image,huang2017arbitrary}, with an emphasis on the texture generation and leaving the geometry untouched. EditNeRF \cite{liu2021editing} proposed to update the latent embeddings of NeRF for editing both shape and appearance. 
ClipNeRF \cite{wang2022clip}, NeRF-Art \cite{wang2023nerf} and Blended-NeRF \cite{gordon2023blended} adopted CLIP \cite{radford2021learning} to maximize the similarity between NeRF and the given text descriptions. DFFs \cite{kobayashi2022decomposing} distilled image feature from DINO \cite{caron2021emerging} and LSeg \cite{li2022languagedriven} into radiance fields and allows direct editing of properties such as shape, size and color. More closely related to our work, InstructNeRF2NeRF \cite{instructnerf2023} adopted the 2D instruction-based editing method InstructPix2Pix \cite{brooks2023instructpix2pix} to update the scene with the Iterative Dataset Update (Iterative DU) technique. Besides iteratively updating the dataset \cite{instructnerf2023,mirzaei2023watch,sun2023instruct}, the iterative loss SDS \cite{poole2022dreamfusion} was adopted by some methods for text-based 3D editing, such as Vox-E \cite{sella2023vox}, FocalDreamer \cite{li2023focaldreamer}, AvatarStudio \cite{AvatarStudio} and DreamEditor \cite{zhuang2023dreameditor}. 
DN2N \cite{fang2023text} accelerates existing editing methods by training a generalizable NeRF so that each editing prompt will not require extra training steps like the above iterative methods.
Our editing method employs a distinct pipeline, which obviates the necessity for iterative updates to edit 3D scenes, as the proposed regularization ensures the multiview consistency of the edited images. Notably, our method exhibits a marked efficiency compared to the aforementioned technique and often finishes within minutes.
Concurrent with our work, Gaussian splatting representation is adopted to accelerate the radiance editing \cite{GaussianEditor,chen2023gaussianeditor}, and we consider their methods orthogonal to us since we speed up the editing on the diffusion side.

\begin{figure*}
    \centering
    \includegraphics[width=\textwidth,trim={1em 0 0 0}]{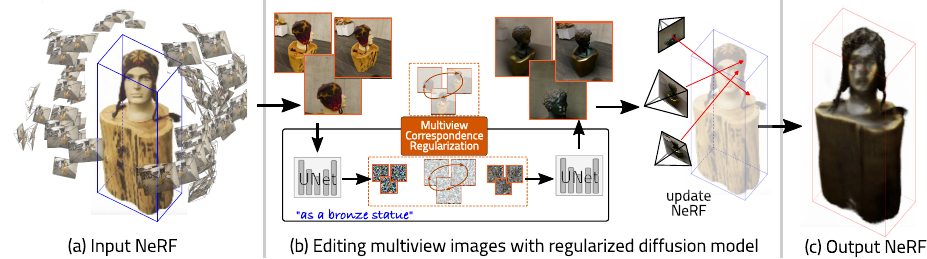}
    \caption{\textbf{The overall framework of multiview correspondence regularized diffusion.} 
    We regularize that the output obtained during the denoising process of diffusion aligns with the input multiview images in terms of multiview correspondence
    }
    \label{fig:framework}
\end{figure*}

\begin{figure}
    \centering
    \includegraphics[width=\columnwidth]{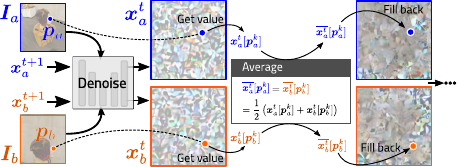}
    \caption{\textbf{Multiview correspondence regularization on the diffusion process.} $\bm{I}_a$ and $\bm{I}_b$ are input views with correspondence pair $(\bm{p}_a,\bm{p}_b)$. $\bm{x}_a^t$ and $\bm{x}_b^t$ are the samples at time $t$ as defined in \cref{eq:gen0}. ``Denoise'' step refers to \cref{eq:diffusion}. 
    }
    \label{fig:framework-average}
\end{figure}

\section{Preliminaries}
\paragraph{Diffusion Models.} 
Diffusion models \cite{song2019generative,ddpm,ddim,croitoru2023diffusion} aim to approximate a given data distribution $q(\bm{x}_0)$ by learning a model distribution $p_{\theta}(\bm{x}_0)$ that is both a good approximation of $q(\bm{x}_0)$ and allows for efficient sampling. A significant contribution in this field is the development of Denoising Diffusion Probabilistic Models (DDPMs \cite{sohl2015deep,ddpm,song2020score}), which belong to latent variable models described by the following form:
\begin{equation}
    p_\theta(\bm{x}_0) = \int p_\theta(\bm{x}_{0:T}) \mathrm{d} \bm{x}_{1:T} \label{eq:gen0}
\end{equation}
where $\bm{x}_1, \ldots, \bm{x}_T$ are latent variables in the same sample space as $\bm{x}_0$ (denoted as $\mathcal{X}$), and $p_\theta(\bm{x}_{0:T}) := p_\theta(\bm{x}_T) \prod_{t=1}^{T} p^{(t)}_\theta(\bm{x}_{t-1} | \bm{x}_t)$. 
In DDPM \cite{ddpm}, we have $p(\bm{x}_T)=\mathcal{N}(\bm{x}_T; \mathbf{0}, \mathbf{I})$ and $p_\theta(\bm{x}_{t-1}|\bm{x}_t)=\mathcal{N}(\bm{x}_{t-1};\mu_\theta(\bm{x}_t,t),\sigma^2_t \mathbf{I})$, where the trainable component $\mu_\theta(\bm{x}_t,t)$ has the time-dependent constant variance $\sigma^2_t$.
The forward process is then defined to learn $\mu_\theta$ as
$q(\bm{x}_{1:T}|\bm{x}_0)=\prod_{t=1}^{T} q(\bm{x}_t|\bm{x}_{t-1}),$
where $q(\bm{x}_t|\bm{x}_{t-1})=\mathcal{N}(\bm{x}_t;\sqrt{1-\beta_t} \bm{x}_{t-1},\beta_t \mathbf{I})$ with $\beta_t$ as predefined constants. Next, in DDPM we have
\begin{equation}
    \mu_\theta(\bm{x}_t,t)=\frac{1}{\sqrt{\alpha}_t}\left(\bm{x}_t - \frac{\beta_t}{\sqrt{1-\bar{\alpha}_t}} \bm{\epsilon}_\theta (\bm{x}_t, t)\right),
    \label{eq:diffusion}
\end{equation}
where $\alpha_t$ and $\bar{\alpha}_t$ are constants derived from $\beta_t$ and $\bm{\epsilon}_\theta$ is a noise predictor, usually parameterized by neural networks.
In image generation tasks, the generation can be conditioned on inputs like text or reference images, and we have 
\begin{equation}
    \bm{\epsilon}^{t}=\bm{\epsilon}_\theta(\bm{x}_t,t,\bm{c}),
    \label{eq:denoise}
\end{equation}
where $\bm{c}$ is the conditioning inputs.

\paragraph{Radiance Fields.}
3D contents in radiance fields are represented by functions that take in spatial location $(x,y,z)$ and viewing direction $(\theta,\phi)$ and output the radiance and occupancy of that location. The radiance value and occupancy for all points along a camera ray are considered to render an image from the radiance fields. 
Volume rendering \cite{mildenhall2020nerf} facilitates the generation of radiance fields by accessing and analyzing all points present on the camera rays. Recent advancements, such as the utilization of 3D Gaussian points \cite{kerbl20233d} to parameterize a radiance field, have led to more efficient rendering pipelines. we opt to generalize the rendering process by symbolically representing it as 
\begin{equation}
    \bm{I}=\pi(\mathcal{F}, \mathbf{P}), \label{eq:render}
\end{equation}
where $\bm{I}$ is the rendered image, $\mathcal{F}$ is the radiance field, and $\mathbf{P}$ is the camera projection matrix (\ie, intrinsics and extrinsics).

\section{Our Method}
The overall framework of our proposed method is straightforward. First, we edit multiview images by the provided textual prompt. Subsequently, we proceed to update the radiance field with these edited images.
Our approach differs from Instruct-NeRF2NeRF in two key respects. Firstly, while Instruct-NeRF2NeRF performs image editing on one image at a time, we apply it to a batch of multiview images. Secondly, we employ a single-pass dataset update scheme, whereas Instruct-NeRF2NeRF requires iterative updates to the dataset.

\subsection{Multiview Correspondence Regularization}
The goal of correspondence regularization is to maintain the correspondence among the multiview images after editing. Our method for pursuing this objective is quite clear: \textit{we aim to ensure that the samples $\bm{x}_{0:T}$ during diffusion are aligned with the inputs from the multiple views during the denoising process}. 

\begin{figure}
\small
    \centering
    \includegraphics[width=0.32\columnwidth]{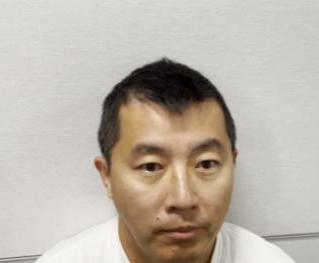}
    \includegraphics[width=0.32\columnwidth]{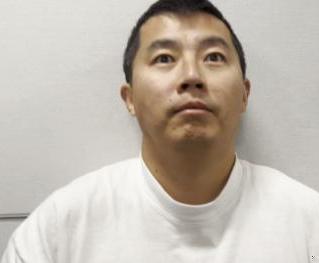}
    \includegraphics[width=0.32\columnwidth]{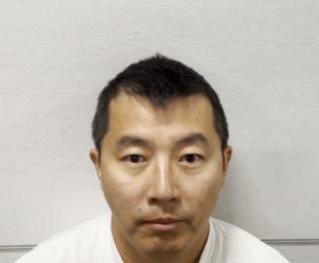}
    \\
    (a) Input views\\
    \includegraphics[width=0.32\columnwidth]{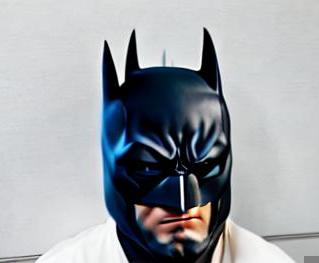}
    \includegraphics[width=0.32\columnwidth]{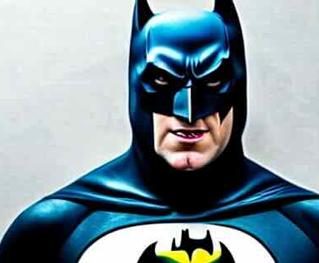}
    \includegraphics[width=0.32\columnwidth]{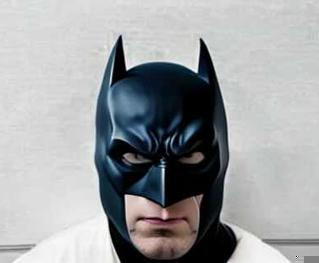}
    \\
    (b) View-by-view editing w/o regularization\\
    \includegraphics[width=0.32\columnwidth]{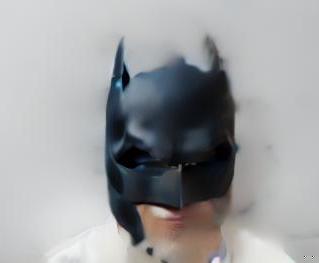}
    \includegraphics[width=0.32\columnwidth]{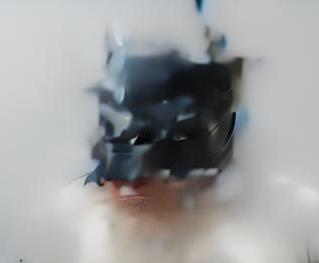}
    \includegraphics[width=0.32\columnwidth]{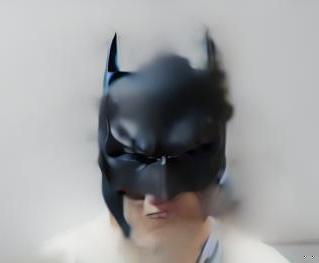}
    \\
    (c) With regularization on all denoising steps (\ie using \cref{eq:averageall})\\
    \includegraphics[width=0.32\columnwidth]{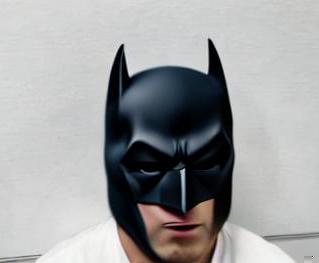}
    \includegraphics[width=0.32\columnwidth]{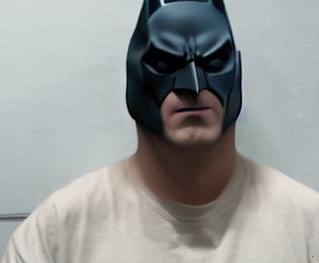}
    \includegraphics[width=0.32\columnwidth]{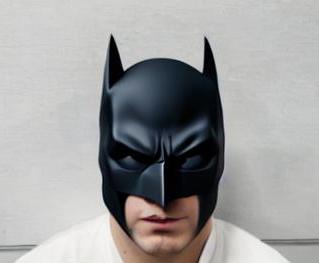}
    \\
    (d) With our softened regularization (\cref{sec:soft-reg})\\
    \caption{\textbf{Multiview image editing results} of (a) input views: (b) view-by-view editing baseline without any regularization; (c) enforcing multiview correspondence on all denoising steps; (c) our softened correspondence regularization. (Prompt: \textit{``Turn him into Batman''}.)}
    \label{fig:blind}
\end{figure}

\begin{figure*}
\small
    \centering
    \includegraphics[width=\textwidth,trim={0 1.5em 0 0}]{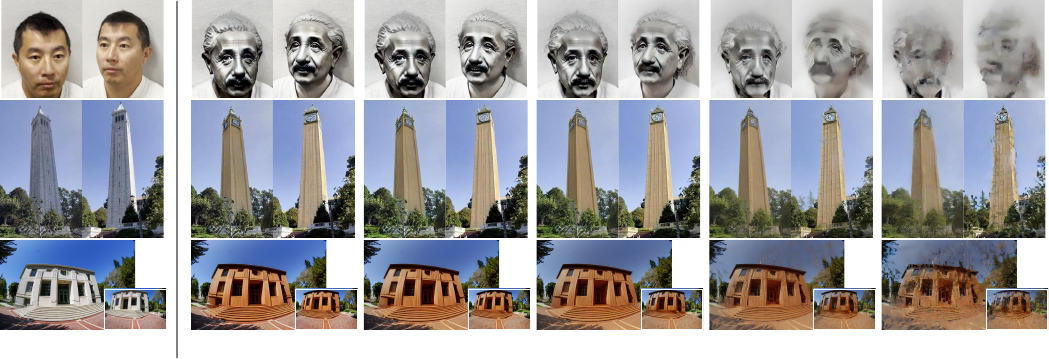}
    \begin{tabular}{
    @{}>{\PreserveBackslash\centering}p{0.17\textwidth}@{}
    @{}>{\PreserveBackslash\centering}p{0.17\textwidth}@{}
    @{}>{\PreserveBackslash\centering}p{0.17\textwidth}@{}
    @{}>{\PreserveBackslash\centering}p{0.17\textwidth}@{}
    @{}>{\PreserveBackslash\centering}p{0.17\textwidth}@{}
    @{}>{\PreserveBackslash\centering}p{0.17\textwidth}@{}}
    Input views&$T_\mathrm{end}=10$&$T_\mathrm{end}=20$&$T_\mathrm{end}=30$&$T_\mathrm{end}=40$&$T_\mathrm{end}=50$\\
    \end{tabular}
    \caption{\textbf{Impact of the regularization ending step $T_\text{end}$}. Moving the regularization to early steps can avoid unwanted blurriness in the generation. (Prompt: \textit{``Turn him into Albert Einstein''}, \textit{``Turn the tower into Big Ben''} and  \textit{``As a wooden building''}.)}
    \label{fig:average-step}
\end{figure*}

Without loss of generality, let us consider two views, denoted as $\bm{I}_a$ and $\bm{I}_b$, and suppose there are $K$ corresponding points denoted as $\{(\bm{p}_a^k, \bm{p}_b^k)\}_{k=1}^K$.
According to the definition of denoising in equation \ref{eq:denoise}, now the conditioning for editing the two images becomes $\bm{c}_a=\{\bm{c}_\text{prompt}, \bm{I}_a\}$ and $\bm{c}_b=\{\bm{c}_\text{prompt}, \bm{I}_b\}$.
For time $t$, we further have the estimated sample $\bm{x}^{t}_a$ and $\bm{x}^{t}_b$ based on the conditioning. 
Our regularization is based on a distance on the estimated noise based on the corresponding points in the input, namely correspondence distance, as defined by
\begin{equation}
    d\left(\bm{x}_a, \bm{x}_a;\{(\bm{p}_a^k, \bm{p}_b^k)\}_{k=1}^K\right) = \frac{1}{K} \sum_{k=1}^K\left(\bm{x}_a[\bm{p}_a^k]-\bm{x}_b[\bm{p}_b^k]\right),
    \label{eq:distance}
\end{equation}
where $\bm{x}^{t}[\bm{p}^k]$ means the value of $\bm{x}^{t}$ on the coordinate $\bm{p}^k$.
Our regularization can be then formulated as a loss for minimizing the distance, \ie,
\begin{equation}
    L_\text{reg}(\bm{x}_a, \bm{x}_b) = \min_{\bm{x}_a, \bm{x}_b} d\left(\bm{x}_a, \bm{x}_a;\{(\bm{p}_a^k, \bm{p}_b^k)\}_{k=1}^K\right). 
    \label{eq:reg}
\end{equation}
Incorporating the regularization loss in the retraining of the diffusion network can indeed be a viable approach. However, it is worth noting that this method may result in significant training time requirements, making it less desirable. Fortunately, there exists a closed-form solution for the regularization loss, that is,
\begin{equation}
    \overline{\bm{x}_a^t}[\bm{p}_a^k]=\overline{\bm{x}_b^t}[\bm{p}_b^k]=\frac{1}{2}\left(\bm{x}^t_a[\bm{p}_a^k]+\bm{x}^t_b[\bm{p}_b^k]\right), \forall t\in[0,T]
    \label{eq:averageall}
\end{equation}
where $\overline{\bm{x}_a^t}$ and $\overline{\bm{x}_b^t}$ are the results with the perfect multiview correspondence. \cref{eq:averageall} can be interpreted that we calculate the mean value of all correspondence pairs on the outputs from \cref{eq:denoise} and continue diffusion with \cref{eq:diffusion}.
However, directly using \cref{eq:averageall} leads to undesired over-blurred image outputs, as shown in \cref{fig:blind}(c).

The occurrence of blurred outcomes can be comprehended. 
This is predominantly due to the presence of noise in the pairs of multiview correspondences (\ie, $\{(\bm{p}_a^k, \bm{p}_b^k)\}_{k=1}^K$). This noise can arise from either imprecise estimation of correspondences or due to the diffusion backbone we have employed, which relies on downsampling in the latent space. Even if the pixel-level correspondence appears to be reliable, it does not guarantee accurate correspondence in the latent space. Furthermore, there may exist instances where strict correspondence is lacking in the latent space. In cases where the correspondence is noisy, the application of regularization will result in a blurred effect, as it produces an averaged outcome across different patches.
We propose to soften the regularization in the following section to address the undesired blurry outputs.

\subsection{Softened Regularization}\label{sec:soft-reg}
The cause of the blurred generation can be attributed to an excessively strong regularization. Thus, it is natural to consider decreasing the strength of regularization in the diffusion process. Our insight is that this reduction in regularization should preserve the majority of the semantics without the necessity of retaining all image details. For example, if we want to edit human facial images while maintaining the multiview alignment of the facial structure is essential, inconsistencies in details such as wrinkles across different views may not be of significant concern.

Drawing on previous research \cite{meng2021sdedit}, which suggests that diffusion models primarily focus on capturing the overall structure during the initial denoising stages and pay more attention to finer details in subsequent stages, we propose to apply \cref{eq:averageall} solely to the early steps of the diffusion process. Formally, we introduce a threshold, denoted as $T_\text{end}$, which replaces $T$ in \cref{eq:averageall}. For denoising steps beyond $T_\text{end}$, no regularization is employed, and the denoising steps are carried out as per the standard procedure outlined in \cref{eq:denoise}.

In Figure 1, we present the visual representation to demonstrate the influence of various selections of $T_{\text{end}}$. The level of indistinctness grows with the increase in $T_{\text{end}}$. Empirically, we found that by setting $T_{\text{end}}$ to 10, we achieve favorable multiview-consistent editings. Consequently, we have employed this particular value for all of our experiments using the InstructPix2Pix \cite{brooks2023instructpix2pix}.

\subsection{Updating Radiance Fields}\label{sec:style}
\paragraph{Style Loss.}
By utilizing the multiview images obtained by regularized diffusion in the previous section, the process of updating the radiance fields becomes straightforward. First, we follow Instruct-NeRF2NeRF and sample a batch of $N$ training cameras $\{\mathbf{P}_n\}_{n=1}^N$. A set of images is first rendered as in \cref{eq:render}, then we apply multiview correspondence regularized 2D image editing on it. Next, with the same training cameras $\{\mathbf{P}_n\}_{n=1}^N$, we randomly render image patches with these cameras, and sample corresponding patches from the generated images in the previous step. Here, we slightly abuse the notation between images and patches, and denote the rendered patches as $\bm{I}_n$ and the corresponding edited patches as $\bm{I}_n^{\mathrm{gen}}$.
However, direct training with $\bm{I}_n^{\mathrm{gen}}$ leads to artifacts, which arise from the presence of inconsistent areas since the regularization improves consistency but is not perfect. Therefore, we propose to leverage a loss function that is less sensitive to multiview inconsistency, and through empirical analysis, we find that the style loss based on the Gram matrix \cite{gatys2016image} is an excellent candidate for this purpose. The total training loss is then
\begin{equation}
    L = \frac{1}{N}\sum_{n=1}^N\left( (\bm{I}_n-\bm{I}_n^{\mathrm{gen}})^2 + \lambda (G(\bm{I}_n)-G(\bm{I}_n^{\mathrm{gen}}))^2 \right),
    \label{eq:loss}
\end{equation}
where $G(\cdot)$ is the Gram matrix \cite{PerceptualLosses} of the input, and $\lambda$ is set to 0.1 in our experiments empirically.

\paragraph{Single-Pass Dataset Update.}
Instruct-NeRF2NeRF proposed a scheme for updating NeRF named iterative dataset update, whereby a randomly selected image from the dataset is edited and subsequently added to the training views. The updated training views are then used to update the NeRF model over multiple steps. In their experimental settings, the NeRF updating step is 10 steps, so the editing model (Instruct-Pix2Pix) is called every 10 steps with a randomly sampled image for updating the training dataset.

We adopt a different strategy for updating NeRF. Thanks to the multiview-consistently edited images, we can directly edit NeRF with those images without the need for iteratively updating the dataset. We denote the set for storing the generated image as $\mathcal{D}_{\mathrm{gen}}$, which is empty at the beginning of the training. Denote the input views as $\mathcal{D}_{\mathrm{input}}=\{\bm{I}_v\}$, we randomly sample a batch of images $\{\bm{I}_b\}_{b=1}^B$ from $\{\bm{I}_v\}$ and then apply the multiview correspondence regularized diffusion on the batch of images. Denote the generated images as $\{\bm{I}^{\mathrm{gen}}_b\}_{b=1}^B$, we add them to the set of generated images, \ie, $\mathcal{D}_{\mathrm{gen}}:=\mathcal{D}_{\mathrm{gen}}\bigcup\{\bm{I}^{\mathrm{gen}}_b\}_{b=1}^B$.  After that, we edit NeRF with $\mathcal{D}_{\mathrm{gen}}$ and training with \cref{eq:loss} for 200 steps. A comparison of Single-Pass Dataset Update and Iterative Dataset Update can be found below.\\
\begin{minipage}{0.49\columnwidth}
\begin{afloat}[H]
\small
\centering
    \caption{}\label{algorithm:iterative}
    \begin{algorithmic}[1]
        \State Init NeRF $\mathcal{F}$ with
               input $\mathcal{D}_{\mathrm{input}}=\{\bm{I}_v\}$
        \State Init \textcolor{blue}{$\mathcal{D}_{\mathrm{gen}}=\mathcal{D}_{\mathrm{input}}$}
        \Repeat
        \State Sample \textcolor{blue}{$\bm{I}_i$}
        \State $\textcolor{blue}{\bm{I}_i^{\mathrm{gen}}}=\text{Edit}(\textcolor{blue}{\bm{I}_i})$
        \State \textcolor{blue}{$\mathcal{D}_{\mathrm{gen}}[i]=\bm{I}_i^{\mathrm{gen}}$}
        \For{$i$ in range(\textcolor{blue}{$N_{\text{IU}}$})}
        \State Update $\mathcal{F}$ with $\mathcal{D}_{\mathrm{gen}}$
        \EndFor
        \Until {reach max steps}
        \State \textbf{Return} $\mathcal{F}$
    \end{algorithmic}
\end{afloat}
\end{minipage}
\begin{minipage}{0.49\columnwidth}
\begin{bfloat}[H]
\small
    \centering
    \caption{}\label{algorithm:single-pass}
    \begin{algorithmic}[1]
        \State Init NeRF $\mathcal{F}$ with
               input $\mathcal{D}_{\mathrm{input}}=\{\bm{I}_v\}$
        \State Init \textcolor{Bittersweet}{$\mathcal{D}_{\mathrm{gen}}=\{\}$}
        \Repeat
        \State Sample \textcolor{Bittersweet}{$\{\bm{I}_b\}_{b=1}^B$}
        \State $\textcolor{Bittersweet}{\{\bm{I}^{\mathrm{gen}}_b\}}=\text{\textcolor{Bittersweet}{Reg}Edit}(\textcolor{Bittersweet}{\{\bm{I}_b\}})$
        \State \textcolor{Bittersweet}{$\mathcal{D}_{\mathrm{gen}}:=\mathcal{D}_{\mathrm{gen}}\bigcup\{\bm{I}^{\mathrm{gen}}_b\}$}
        \For{$i$ in range(\textcolor{Bittersweet}{$N_{\text{SU}}$})}
        \State Update $\mathcal{F}$ with $\mathcal{D}_{\mathrm{gen}}$
        \EndFor
        \Until {reach max steps}
        \State \textbf{Return} $\mathcal{F}$
    \end{algorithmic}
\end{bfloat}
\end{minipage}

\vspace{5pt}
\noindent
\textcolor{blue}{$N_{\text{IU}}$} and \textcolor{Bittersweet}{$N_{\text{SU}}$} are tuneable hyper-parameters, with default values set at 10 (in \cite{instructnerf2023}) and 200, respectively.

\begin{figure*}
\small
    \centering
    \includegraphics[width=\textwidth,trim={0 0.8em 0 0}]{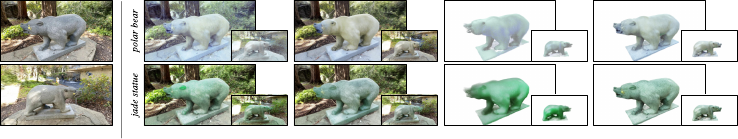}\\
    \begin{tabular}{
    @{}>{\PreserveBackslash\centering}p{0.15\textwidth}@{}
    @{}>{\PreserveBackslash\centering}p{0.26\textwidth}@{}
    @{}>{\PreserveBackslash\centering}p{0.19\textwidth}@{}
    @{}>{\PreserveBackslash\centering}p{0.21\textwidth}@{}
    @{}>{\PreserveBackslash\centering}p{0.19\textwidth}@{}}
    Input NeRF&Ours (2 mins)&I-N2N \cite{instructnerf2023} (20 mins)&
    Vox-E \cite{sella2023vox} \textcolor{gray}{(55 mins)}&
    NeRF-Art \cite{wang2023nerf} \textcolor{gray}{(2 hours)}\\
    \end{tabular}
    \caption{\textbf{Comparison with SoTA methods.} For ours and Instruct-NeRF2NeRF, we use instructional text furnished with the suffix \textit{``Turn the bear into a ...''.} Background of input images is removed for Vox-E \cite{sella2023vox} and NeRF-Art \cite{wang2023nerf}. }
    \label{fig:sota-compare}
\end{figure*}

\begin{figure*}
\small
    \centering
    \includegraphics[width=\textwidth,trim={0 1.5em 0 0}]{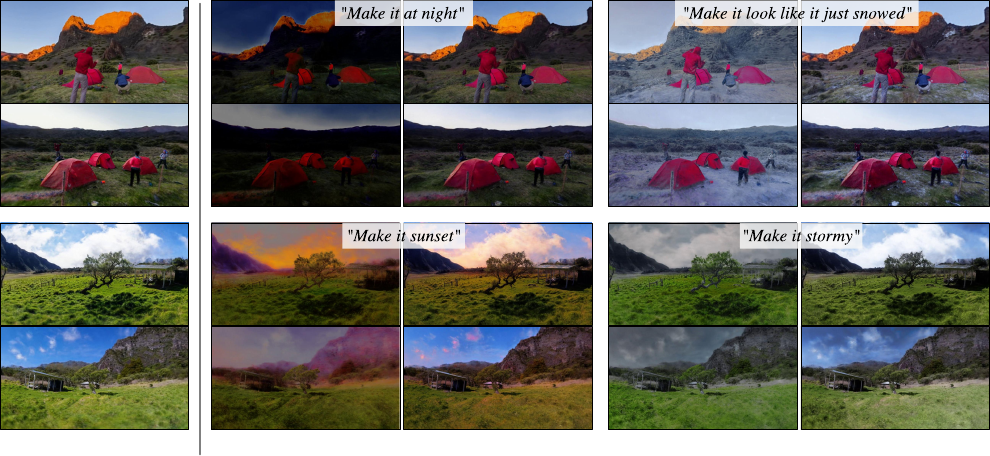}\\
    \begin{tabular}{
    @{}>{\PreserveBackslash\centering}p{0.2\textwidth}@{}
    @{}>{\PreserveBackslash\centering}p{0.21\textwidth}@{}
    @{}>{\PreserveBackslash\centering}p{0.19\textwidth}@{}
    @{}>{\PreserveBackslash\centering}p{0.21\textwidth}@{}
    @{}>{\PreserveBackslash\centering}p{0.19\textwidth}@{}}
    Input NeRF&Ours (30s)&I-N2N \cite{instructnerf2023} (20 mins)&Ours (30s)&I-N2N \cite{instructnerf2023} (20 mins)\\
    \end{tabular}
    \caption{\textbf{Comparison of the NeRF editing efficiency.} Our method can effectively edit NeRFs in just 30 seconds for scene styles.}
    \label{fig:sota-step-compare}
    \vspace{-1em}
\end{figure*}

\begin{figure}
    \centering
    \includegraphics[width=\columnwidth,trim={0cm 0.3cm 0cm 0cm},clip]{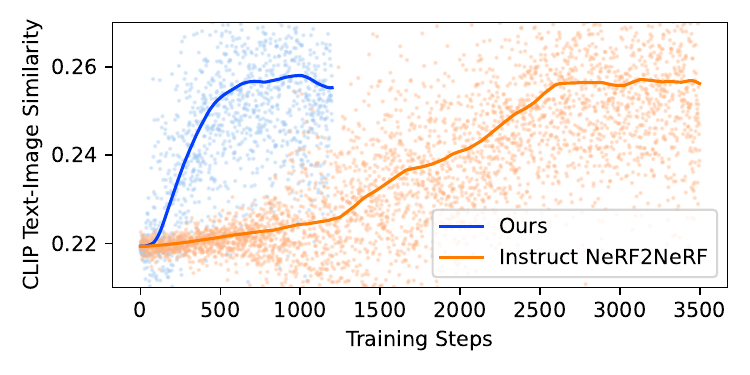}
    \caption{\textbf{CLIP Text-Image Similarity score during training.} We repeat the experiments 10 times as the variance of the CLIP Text-Image Similarity score is large. The exponential moving average is adopted to smooth the mean value curve.}
    \label{fig:clip-similarity}
\end{figure}

\begin{figure*}
    \centering
    \includegraphics[width=\textwidth]{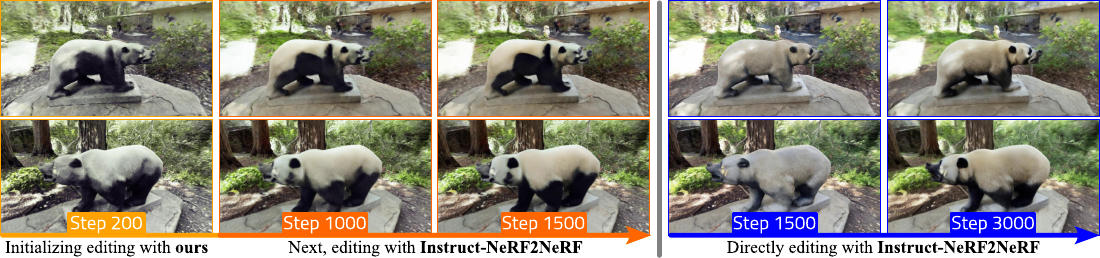}
    \caption{\textbf{Our method can be combined with Instruct-NeRF2NeRF.} Initializing with our method can speed up the editing with on-par performance. (Prompt: \textit{``Turn the bear into a panda.''})}
    \label{fig:compatible}
\end{figure*}

\begin{figure}
    \centering
    \includegraphics[width=\columnwidth]{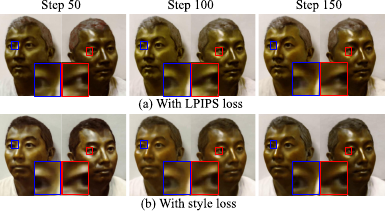}
    \caption{\textbf{Comparison of training with LPIPS loss and style loss.} LPIPS loss is adopted in Instruct-NeRF2NeRF. (Prompt: \textit{``As a bronze bust.''})}
    \label{fig:lpips-style-compare}
    \vspace{-1em}
\end{figure}

\section{Experiments}
Our experiments were conducted following the framework proposed in Instruct-NeRF2NeRF \cite{instructnerf2023}. Specifically, we trained the input NeRF model using the Nerfacto model within the Nerfstudio \cite{nerfstudio}. To evaluate the performance of our method, we conducted experiments on the dataset provided by Instruct-NeRF2NeRF, with the exception that scenes containing identifiable human faces were replaced with our re-captured versions for privacy and legal considerations.

Our approach significantly enhances the existing method, Instruct-NeRF2NeRF \cite{instructnerf2023}, concerning its efficacy in editing. However, illustrating the efficiency effects of editing at various optimization steps is not straightforward through images. Therefore, we earnestly recommend that readers refer to our supplementary video for a more comprehensive comparison of the editing efficiency. 
Time in this paper is all measured on a single A100. 
For our method, editing a batch of 4 multiview images with our regularization takes around 30 seconds, and 1 step of NeRF optimization takes around 0.15 seconds.

\subsection{Comparison with State-of-The-Arts}
We primarily compare our methodology against the state-of-the-art method Instruct-NeRF2NeRF \cite{instructnerf2023}.
We begin by showing the CLIP Text-Image Similarity scores of the edited NeRF along with training iterations, as depicted in \cref{fig:clip-similarity}. The CLIP Text-Image Similarity score represents the cosine similarity between the textual embedding and the image embedding derived from CLIP. We use the text about the target object within the given instructions. For instance, if the instruction specifies "Turn the bear into a grizzly bear," we utilize "a grizzly bear" as the text input for computing the similarity scores. Notably, we observe that the scores exhibit a high variance, with significant fluctuations occurring after each iteration. Consequently, we repeat the experiment 10 times and present the smoothed average value for analysis.
We can observe that our proposed approach exhibits significantly accelerated convergence in comparison to the Instruct-NeRF2NeRF method, all the while achieving comparable performance outcomes following optimization completion.

Furthermore, in \cref{fig:sota-compare}, we present a visual comparison between our method and Instruct-NeRF2NeRF, Vox-E \cite{sella2023vox} and NeRF-Art \cite{wang2023nerf}. 
While NeRF-Art \cite{wang2023nerf} and Vox-E \cite{sella2023vox} have shown commendable performance, we have opted not to report their time metrics. This is due to the 3D representation adopted in their methodology, which makes it hard to offer a justifiable comparison. 
We showcase the efficacy of our approach in \cref{fig:sota-step-compare}. Remarkably, we observe that our method excels in editing scene style, potentially owing to the ability of our regularization to attain superior multiview coherence within this particular context.

\subsection{Ablation Studies}
\paragraph{Compatibility to Instruct-NeRF2NeRF.}
We have demonstrated the effectiveness of our approach in the preceding experimental section. However, it should be noted that the final editing outcomes may not always match the level of quality achieved by Instruct-NeRF2NeRF. Although the similarity scores depicted in \cref{fig:clip-similarity} indicate a close performance between the two methods, in practice, we have observed that our approach and Instruct-NeRF2NeRF exhibit different generation patterns. In certain instances, it becomes challenging to unequivocally assert that our method is consistently more visually appealing than Instruct-NeRF2NeRF. 

To address this issue, we propose a solution that involves combining our method with Instruct-NeRF2NeRF. By initiating the optimization process with our approach and subsequently switching to Instruct-NeRF2NeRF, we can achieve improved results. The outcomes of this combined method are presented in \cref{fig:compatible}. It is evident from the figure that following initialization with our method, Instruct-NeRF2NeRF requires only 800 additional steps to produce outcomes of comparable quality to those achieved with 3000 steps without initialization.
\paragraph{Style Loss.}
In \cref{sec:style}, we claimed that incorporating style loss can effectively mitigate the influence of inconsistency in the input edited views. In \cref{fig:lpips-style-compare}, we present a comparative analysis between the utilization of style loss and the default LPIPS loss in Instruct-NeRF2NeRF. Remarkably, we observe that the images exhibit enhanced sharpness and a more consistent alteration in style throughout the training procedure. As an illustration, we note the gradual transition to a slightly green hue on the faces during the middle stages of training with LPIPS loss.

\section{Conclusion}
We introduce an efficient framework for editing NeRF. Our framework entails the editing of multiple views of images using the proposed multiview correspondence regularization. Subsequently, we perform optimization of NeRF using these manipulated images. Our approach demonstrates considerable efficiency when compared to recent SoTA methods such as Instruct-NeRF2NeRF. Unlike these methods, which rely on single-view editing and iterative optimization of NeRF, our approach achieves multiview consistency in the edited images without requiring iterative processes.

\section*{Acknowledgement}
We thank Ayaan Haque for reviewing and verifying the claims presented in this paper.

{
    \small
    \bibliographystyle{ieeenat_fullname}
    \bibliography{main,nerf}
}

\end{document}